\documentclass[10pt,twocolumn,letterpaper]{article}

\usepackage{btas}
\usepackage{times}
\usepackage{epsfig}
\usepackage{graphicx}
\usepackage{amsmath}
\usepackage{amssymb}

\usepackage[usenames,dvipsnames]{xcolor}
\usepackage[normalem]{ulem}
\usepackage{setspace}
\usepackage{times}
\usepackage{epsfig}
\usepackage{amsmath}
\usepackage{amssymb}
\usepackage{makecell}
\usepackage{verbatim}
\usepackage{balance}
\usepackage{color}
\usepackage{xspace}
\usepackage{enumerate}
\usepackage{graphicx}
\usepackage[noend]{algpseudocode}
\usepackage{algorithm}
\usepackage{wrapfig}
\usepackage{microtype}
\usepackage{pifont}
\usepackage{caption}
\usepackage{multirow}
\usepackage{cite}

\usepackage[utf8]{inputenc} 
\usepackage[T1]{fontenc} 
\usepackage{csquotes}

\newcommand{\ignore}[1]{}
\usepackage{fancyhdr}
\usepackage[hyphens]{url}
\usepackage{hyperref}
\btasfinalcopy 

\ifbtasfinal\fi
\begin{document}

	\title{ThirdEye: Triplet Based Iris Recognition without Normalization}

\author{Sohaib Ahmad\\
University of Connecticut\\
Storrs\\
{\tt\small sohaib.ahmad@uconn.edu}
\and
Benjamin Fuller\\
University of Connecticut\\
Storrs\\
{\tt\small benjamin.fuller@uconn.edu}
}

\graphicspath{ {images/} }

\def\shownotes{1}
\ifnum\shownotes=1
\newcommand{\authnote}[2]{{\textcolor{red}{\textsf{#1 notes: }\textcolor{blue}{ #2}}\marginpar{\textcolor{red}{\textbf{!!!!!}}}}}
\else
\newcommand{\authnote}[2]{}
\fi
\newcommand{\bnote}[1]{{\authnote{Ben}{#1}}}
\newcommand{\snote}[1]{{\authnote{Sohaib}{#1}}}

\newcommand{\cmark}{\normalsize \ding{51}}
\newcommand{\xmark}{\ding{55}}

\maketitle
\thispagestyle{empty}

\begin{abstract}
Most iris recognition pipelines involve three stages: segmenting into iris/non-iris pixels, normalization the iris region to a fixed area, and extracting relevant features for comparison.  Given recent advances in deep learning, it is prudent to ask which stages are required for accurate iris recognition.  Lojez et al. (IWBF 2019) recently concluded that the segmentation stage is still crucial for good accuracy.  We ask if normalization is beneficial?

Towards answering this question, we develop a new iris recognition system called ThirdEye based on triplet convolutional neural networks (Schroff et al., ICCV 2015).  ThirdEye directly uses segmented images without normalization.  

We observe  equal error rates of 1.32\%, 9.20\%, and 0.59\% on the ND-0405, UbirisV2, and IITD datasets respectively.  For IITD, the most constrained dataset, this improves on the best prior work.  However, for ND-0405 and UbirisV2, our equal error rate is slightly worse than prior systems.  Our concluding hypothesis is that normalization is more important for less constrained environments.
\end{abstract}


\section{Introduction}
\label{sec:intro}

This work investigates iris biometric~\cite{daugman2004iris,raja2014smartphone,raja2015smartphone}.  Specifically, we focus on \emph{iris recognition}: systems that create an initial \emph{template} of a person and can identify if a \emph{target} corresponds to a previously known identity. Current performance is strong on datasets but lacks when the environment changes.  This is particularly true for learning-based techniques which often specialize to the training dataset. 
Iris recognition systems have three core components:
\begin{enumerate}
    \item Segmentation which identifies pixels of an image as iris or non-iris.
    \item Normalization which maps iris pixels to a fixed dimension array.  This step often transforms from polar to Cartesian coordinates.
    \item Feature extraction which produces a succinct vector that identifies the image.  For different images of the same iris this vector should be stable.  Simultaneously, this vector should vary widely for images of different individuals.
\end{enumerate}
Daugman's seminal work~\cite{daugman2004iris,daugman2009iris} followed this pipeline and used Gabor filters to extract features from a normalized vector.  Matching between the \emph{template} and the \emph{target} was done by computing the Hamming distance between the two vectors.  Wildes~\cite{wildes1997iris}, followed by Ma~\cite{ma2004efficient}, explored  other system-level considerations such as matching efficiency.  In the twenty years since these pioneering works, there has been substantial work to create accurate systems.  In this work we focus on approaches using deep learning (we compare our results to all state of the art approaches).  

Deep learning~\cite{lecun2015deep} uses complex models to learn representations from high dimensional data allowing learning of features from complex datasets. Deep learning has been applied to various biometrics~\cite{liu2016deepiris,wen2016discriminative,park2009periocular,liu2010finger}.
Like prior work~\cite{minaee2016experimental,proencca2018deep,liu2016deepiris,nalla2017toward} we use \emph{convolutional neural networks} or CNN to extract features from the iris.  We defer further discussion of related work until after introducing our system.

The main goal of this work is to understand how much of the above pipeline is necessary for modern learning approaches.  Recent work of Lojez et al.~\cite{lozej2019influence} concluded a separate segmentation stage is still critical to accurate iris recognition.
The question of this work is: 
\begin{displayquote}
Is normalization necessary for accurate iris recognition using modern techniques?
\end{displayquote}

\subsection{Our Contribution}
Our main contribution is a new iris recognition system we call \emph{ThirdEye}.  The core of ThirdEye is a CNN trained using the triplet based training framework~\cite{weinberger2009distance}.  This network directly uses a segmented iris image but does not include any normalization.  Triplet based networks train on three input images at a time, two from the same class/iris and the third from a different class/iris.  These three images are known as $\mathtt{target}, \mathtt{pos}$, and $\mathtt{neg}$ respectively.  The current weights are evaluated based on the objective:
\[
d(\mathtt{target}, \mathtt{pos}) - d(\mathtt{target}, \mathtt{neg}).
\]
Triplet based networks focus simultaneously on minimizing distance between images in the same class and maximizing distance between images in different classes.  Our system is trained in batches; triples are selected (to use for weight updates) by finding the current \emph{hardest} triple.  For each $\mathtt{target}$, the hardest triplet in a batch, the positive instance, $\mathtt{pos}$, is the image in the same class as $\mathtt{target}$ with the maximal distance.  Similarly, the negative instance, $\mathtt{neg}$, is the image not in $\mathtt{target}$ class with minimum distance to $\mathtt{target}$. This process is called triplet mining and is done for every $\mathtt{target}$ in the batch. However, we find triplet networks collapse due to the uniform black regions representing the sclera and pupil.  These regions make all images very similar to a randomly initialized network.  A normalized network would not suffer from this problem.

To avoid this problem, our network begins with a partially trained version of ResNet~\cite{he2016deep}. The network is pre-trained using softmax classification and then trained using triplet loss. For generalization purposes, the training and testing sets inside a datasets are completely disjoint (have different classes).  Lastly, we add several test time augmentations to produce an output feature vector (see Section~\ref{sec:design}).  

In this work we consider three datasets: 
\begin{enumerate}
    \item  ND-0405~\cite{bowyer2016nd,phillips2008iris}, is a large near infrared (NIR) dataset,
    \item IITD~\cite{kumar2010comparison}, is a more controlled NIR dataset and
    \item UbirisV2~\cite{proenca2010ubiris}, is a visible light and unconstrained dataset. 
\end{enumerate} We report on ThirdEye's accuracy when testing and training on the each dataset.  We also show our system has promise for generalization by training on the ND-0405 dataset and testing on the UbirisV2 and IITD datasets.  EERs (equal error rate) of our system and state of the art are described in Table~\ref{tab:summary}.

\begin{table}
\centering
\small
\begin{tabular}{l | r | r | l | r}
     Dataset &  ERR & C. EER &\multicolumn{2}{c}{Prior best}   \\\hline
     ND-0405& 1.32\% & - & Zhao et al.~\cite{zhao2017towards} & .99\%\\
     UbirisV2 & 9.20\% & 35.00\% & Gangwar et al.~\cite{gangwar2019deepirisnet2} & 8.50\%\\
     IITD & 0.59\% & 1.90\% & Zhao et al.~\cite{zhao2017towards} & 0.64\%
\end{tabular}
\caption{Summary of accuracy results. Full results in are Tables~\ref{tab:ND-0405},~\ref{tab:ubiris}, and~\ref{tab:IITD}. The C. EER column describes the accuracy of ThirdEye when training on ND-0405 and testing on one of the other datasets.}
\label{tab:summary}
\end{table}

Our accuracy rates are competitive with state of the art systems that use normalization.  The accuracy is achieved under coarse segmentation~\cite{hofbauer2014ground} which does not remove eye lash occlusions and reflections\footnote{The framework is thus robust under imperfect segmentation, notably on the Ubiris.v2 dataset.}. Better segmentation would increase recognition accuracy with modern methods. The works of~\cite{kinnison2019learning} and~\cite{kerrigan2019iris} explore the link between segmentation and correct recognition. 

Recently, Lozej et al. who argued that normalization on harder datasets might be erroneous~\cite{lozej2019influence} propagating to recognition errors. However, the most constrained dataset (IITD) is where our accuracy improves over state of the art. Accuracy rates for a slightly harder NIR dataset (ND-0405) are less than state of the art. The least constrained dataset (Ubiris) is where our accuracy is furthest from state of the art.  Thus, our current hypothesis is normalization is least useful in constrained environments but is still important in unconstrained environments.    We discuss this further in Section~\ref{sec:results}.

\subsection{Related Work}
Before turning to the technical description of our design we provide more background on prior work. Discussing all prior work is not possible in this space.  We focus on the major innovations in iris recognition and then turn to learning based approaches relevant to our work. The seminal work of Daugman~\cite{daugman2009iris} proposed using Gabor filters on normalized iris images. The output from convolving the iris image with the Gabor filter yields an iriscode. These iriscodes are compared using Hamming distances. Traditional iris recognition algorithms using Gabor filters are reproducible with open source computer vision libraries.  OSIRIS is an open source implementation~\cite{othman2016osiris} that closely follows Daugman's techniques.

Early works rely on iris images captured using specialized cameras having NIR  sensors. Images captured using these sensors expose the intricate details of the iris and are easier to segment and extract features from. With the advent of consumer digital cameras with RGB sensors, iris recognition began including RGB images. The Ubiris dataset~\cite{proencca2005ubiris} followed by the UbirisV2 dataset~\cite{proenca2010ubiris} are collected using RGB sensors.  UbirisV2 incorporates iris at a distance (IAAD) images. More than 90\% of its images include occlusions, are blurred, or off-angle. Due to these factors segmentation, feature extraction, and recognition are all harder. The UbirisV2 dataset is still a challenge for iris recognition algorithms with EER rates higher than EER rates on NIR datasets~\cite{tan2014accurate}. 

 These harder datasets necessitated new feature extraction.  Alonso-Fernandez et al. used scale invariant feature transform (SIFT) to recognize irises~\cite{alonso2009iris}. Tan et al.~\cite{tan2014accurate} use Zernike moments for iris recognition on the Ubiris.v2 dataset. Research work then transitioned to machine learning, an early paper by Raja et al.~\cite{raja2014smartphone} made use of learnt filters for smartphone based iris recognition. These methods improved accuracy over Gabor filters.
 
Recent work used deep learning as it became the mainstay in image recognition. Convolutional neural networks (CNNs) quickly became best in class on ImageNet~\cite{krizhevsky2012imagenet} dataset. CNNs serve as the backbone of classification algorithms due to their strong feature extraction capabilities. CNNs are also popular in iris recognition~\cite{minaee2016experimental,proencca2018deep,liu2016deepiris,nalla2017toward,gangwar2019deepirisnet2,zhao2017towards}. Gangwar et al.~\cite{gangwar2016deepirisnet} introduce a system called DeepIrisNet that uses a CNN for iris recognition.  They also study the robustness of the system under rotations and different splits of the ND-0405 dataset~\cite{bowyer2016nd}. The potential of CNNs for feature extraction is further shown by Nguyen et al.~\cite{nguyen2018iris}, they used a CNN pre-trained on the ImageNet~\cite{deng2009imagenet} dataset.  Rather than training the full CNN for classification they trained a support vector machine (SVM) on the output features of the CNN for iris recognition. Deep learning has allowed accuracy figures to exceed those from traditional methods. To this end deep learning has also been used for iris segmentation~\cite{arsalan2018irisdensenet,ahmad2018unconstrained}, however difficult datasets still pose a problem~\cite{proenca2010iris}.

Our work uses a network inspired from facial recognition networks which introduced the triplet network~\cite{schroff2015facenet} configuration. As mentioned above the core of the triplet approach is to provide three images as input to the network, two from the same class and a third from a different class. The goal of the loss function is to minimize the distance between the two images from the same class while maximizing the distance between images in different classes.  Zhao et al.~\cite{zhao2017towards} use a triplet based CNNs for feature extraction for iris.  Interestingly, they also train a mask net for categorizing iris pixels in the training of the feature extraction network. This work points out the popularity of the iriscode representation and tries to ``reproduce'' a learning pipeline similar to traditional iris code.  Since the goal of our work is to understand what parts of this pipeline are necessary, our techniques naturally differ.

\paragraph*{Organization}
The rest of this work is organized as follows, Section~\ref{sec:design} describes our overall design of the framework, focusing on the feature extraction network, Section~\ref{sec:evaluation} describes the evaluation methodology and the datasets used, Section~\ref{sec:results} discusses the results on the used datasets and Section~\ref{sec:conclusion} concludes.

\section{Design}
\label{sec:design}
ThirdEye's core is a triplet based CNN.  The main distinguishing feature of a CNN is the use of convolutions as internal neurons.  Our starting point is the ResNet-50 architecture introduced by He et al.~\cite{he2016deep}. ResNet-50 is the most widely used benchmark network. Performance comparisons with this network is easy since ImageNet accuracy improvements always compare to the ResNet-50, judging new architecture performance can be interpolated from ImageNet performance of the ResNet-50. We now go over a brief overview of ResNet-50 before describing our major modifications.  

\begin{figure}[t]
    \centering
    \includegraphics[scale = 0.5]{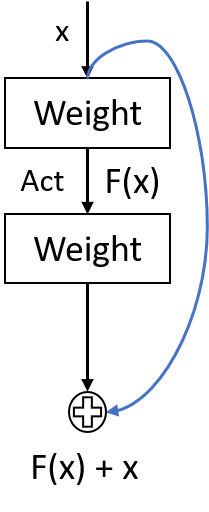}
    \caption{The building block of a ResNet. The blue line shows a skip connection.}
    \label{fig:skipConnection}
\end{figure}

\paragraph{ResNet-50}  In many architectures every neuron at layer $\ell$ is connected to each neuron at layer $\ell+1$.  The main innovation of ResNet-50 is the use of Skip connections which connect neurons at layer $\ell$ with both layer $\ell+1$ and layer $\ell+2$ shown in Figure~\ref{fig:skipConnection}.  Neural networks suffer from accuracy saturation with increasing depth. As the depth increases, less gradient naturally flows through the network, slowing training. Skip connections allow more gradient to flow by adding feature maps of a preceding layer to the current layer ultimately allowing training of deeper networks.

The ResNet-50 has 50 layers divided into five blocks. These blocks are shown in Table~\ref{tab:arch}. After convolutions the final image size is reduced to 7x7. The final feature maps are then passed through an AveragePool layer.  The output of the AveragePool then serves as input to a last fully connected layer. The Softmax classifier is a fully connected layer with neurons equal to the number of classes, taking features from the previous fully connected layer as input. This Softmax layer is used to pre-train the network on the respective iris datasets.  We use the output of this training as our base network. For further training, we take the output of the fully connected layer as a feature vector for the triplet loss function which we describe next.

\begin{table}[t]
	\centering
		
		\begin{tabular}{|l|c|c|}
			
			\hline
			Layer&OutputSize&Kernels\\
			\hline
			Conv1&$128 \times 128$& \makecell{$7\times 7$, 64, stride2 \\$3\times 3$, MaxPool, stride2}  \\
			\hline
			Conv2&$64\times 64$& $\left[\makecell{ 1x1,64\\3x3,64\\1x1,256}\right]$ \large{$\times 3$}  \\
						\hline
			Conv3&$32\times 32$& $\left[\makecell{ 1x1,128\\3x3,128\\1x1,512}\right]$ \large{$\times 4$}  \\
						\hline
			Conv4&$16 \times 16$& $\left[\makecell{ 1x1,256\\3x3,256\\1x1,1024}\right]$ \large{$\times 6$}  \\
									\hline
			Conv5&$8\times 8$& $\left[\makecell{ 1x1,512\\3x3,512\\1x1,2048}\right]$ \large{$\times 3$}  \\
												\hline
			FC&$1\times 1$& \makecell{ AveragePool, 2048-d fc}  \\
			\hline
	\end{tabular} 
	    
\caption{ResNet-50 architecture with an input image size of 256x256}
\label{tab:arch}
\end{table}

\subsection{Triplet loss}
During a second stage of training, we replace the Softmax classifier with a triplet loss function.
A triplet network has the same network duplicated three times with the same weights and takes as input three images or one triplet.  These three images are known as the \emph{anchor image}, $x_a$, the positive image, $x_p$, and the negative image $x_n$.  The anchor image $x_a$ and the positive image $x_p$ should be from the same class while $x_n$ should be from a different class. The triplet loss function is~\cite{schroff2015facenet}:

\begin{equation}
\sum_{i} \left(d(x_{a,i},x_{p,i})    -    d(x_{a,i},x_{n,i})  + \alpha\right)
\end{equation}

The loss minimizes the distance between the anchor and the positive sample while maximizing the distance between the anchor and the negative sample. The distance being minimized is the L2 Euclidean distance. The hyper-parameter $\alpha$ is the separation between like and unlike samples the network is optimizing towards. The value of the margin is dataset dependent and is selected after calculating the accuracy across many values of the margin.  

\begin{figure}[h]
    \centering
    \includegraphics[scale = 0.7]{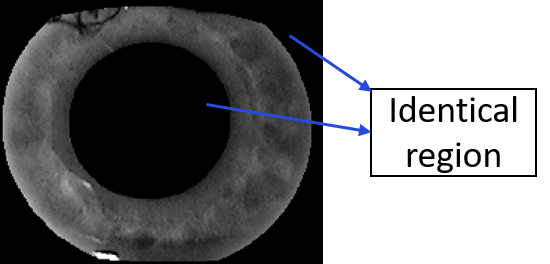}
    \caption{Segmented iris image. A vanilla triplet network collapses at the start of training. The randomly initialized network finds all images similar due to the identical black region. }
    \label{fig:collapsing}
\end{figure}

However, straightforward application of the triplet loss function did not work in our initial experiments without a normalization stage.  To explain why requires a reminder of the type of images used as input to our network.  The inputs are segmented iris images (see Figure~\ref{fig:collapsing}).  The segmentation framework sets non-iris pixels to black.  This region includes the sclera and the pupil.  This behavior is the same for all classes.    Our original triplet network found all images similar and was unable to meaningfully recognize irises.  (If a normalization stage were present, these pixels would be removed before the feature extractor.)

To overcome this problem we employ two techniques. Firstly we use a network that has been pre-trained on the same iris dataset using the softmax classifier. This is done to tune the resulting output distributions of the image embeddings and serve as a starting point to the triplet loss trained network. After pre-training the network is at least 90\% accurate on the training set. 

Secondly we train the network on triplets that are the most difficult first.  This is known as a  Batch hard~\cite{hermans2017defense} triplet loss which explicitly picks \textit{hard} triplets for training. A triplet in the batch hard loss function consists of $x_a, x_p, x_n$ as before.  However, $x_p$ is chosen as the current worst positive instance, that is suppose that $x_{a,i}$ is in class $\mathcal{I}$, 
\[
x_{p,i}=\texttt{argmax}_{x\in\mathcal{I},x\neq x_{a,i}}\,\, d(x_{a,i}, x).
\] 
The negative image is chosen as the current worst negative instance, that is the image that minimizes the distance:
\[
x_{n,i} = \texttt{argmin}_{x\not\in\mathcal{I}}\,\, d(x_{a,i}, x).
\]

\begin{figure}[t]
    \centering
    \includegraphics[scale = 0.70]{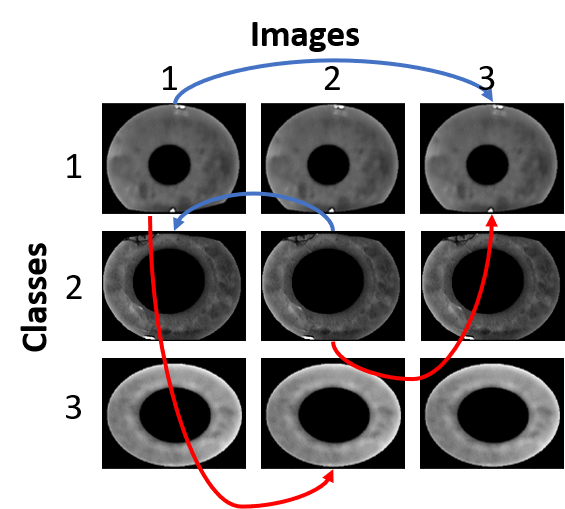}
    \caption{Triplet generation. For two arbitrary targets, hardest positives and hardest negatives are shown denoted by blue and red arrows respectively.}
    \label{fig:tripleGen}
\end{figure}

The triplet mining procedure is explained in Figure~\ref{fig:tripleGen} where two images (row1,col1/row2,col2) serves as our targets. The hardest positive sample for the first target image is the third image in its own class while the hardest negative sample is the second image in the third class. The target combined with the positive and negative sample forms a triple. Every image will serve as a target for tripletformation. In total for the example in Figure~\ref{fig:tripleGen} there will be 9 triples formed. 

A valid triplet is one where there are two positives and a negative. We then distinguish between easy triplet and hard triplets.  A hard triplet is one where:
\[
d(x_{a,i}, x_{p,i}) + m \ge  d(x_{a,i}, x_{n,i}).
\]
 
In the above $m$ is the margin between positive and negative instances.  
The goal of the batch hard loss is to use hard and valid triplets. Thus, overall batch hard triplet loss function is :

\begin{equation}
\begin{split}
TL_{BH}  = \sum_{i,i \text{ is hard} }\left(m +d(x_{a,i}, x_{p,i})- d(x_{a,i}, x_{n,i})\right).
\end{split}
\end{equation}

In the above $x_{a,i}$ and $x_{p,i}$ are the worst images as described above. 

Triplet mining can be done in an offline or online fashion. Offline triplet mining involves calculating embeddings on \textbf{all} images in the training set and training on mined valid triplets from the calculated embeddings. This introduces a large overhead before every epoch of training. Online triplet mining involves calculating embeddings on the fly given a batch of images and training on mined triplets from within the batch. We use the online triplet scheme for training our network due to its efficiency in triplet selection. Instead of finding the worst triples for the entire dataset, we split the dataset into batches that contain a fixed number of classes (which we call $P$) and images per class (which we call $K$) and restrict the search for triplets to within a batch. The online triplet model uses a single model for training and embedding calculation.

We recall that only hard triplets are used to update weights so defining the margin effects accuracy.  However, we use the soft-margin described by Hermans et al.~\cite{hermans2017defense} where an explicit margin is not required. The final loss is calculated as:
\[
TL_{BH} = \log(1+\exp\left(d(x_{a,i},x_{p,i}) - d(x_{a,i},x_{p,i})\right).
\]
In the above $x_{p,i}$ and $x_{n,i}$ are the hardest images as described before.

ThirdEye is trained using un-normalized images as inputs. The segmented images are are \textit{coarsely normalized}, the iris region is in the center of the image but not aligned. We force segmented images from a dataset to have the same resolution through zero padding. The aspect ratio is chosen to be 1:1.  This is because ResNet-50 is tuned to this aspect ratio which is also used with the Imagenet images.  Specific resolutions were chosen per dataset, small enough for ease of training in enabling large enough batch sizes, large enough so that resizing does not introduce artifacts or decrease image quality and are stated in Section~\ref{sec:evaluation}.

\begin{figure*}[t]
    \begin{center}
    \includegraphics[scale = 0.52]{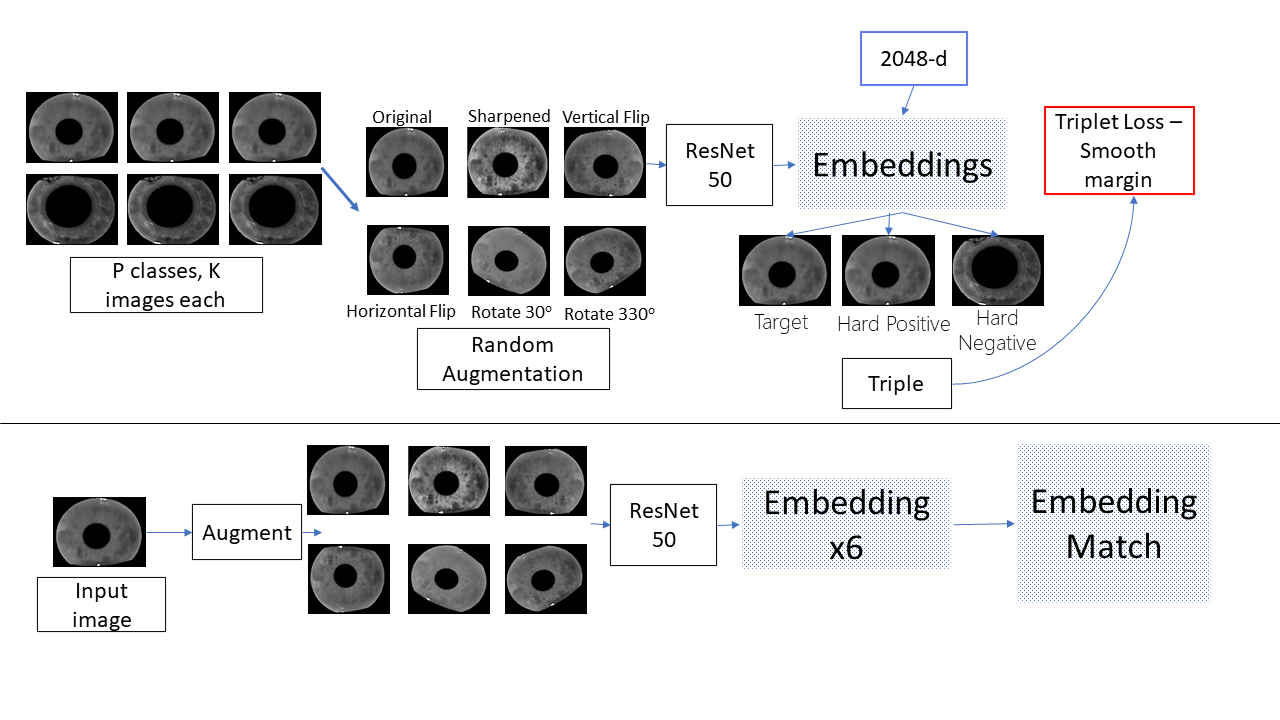}
    \caption{Triplet design. The top part of the figure shows training the ResNet with hard triples. The bottom shows feature vector generation at test time. }
    \label{fig:triplet}

    \end{center}
\end{figure*}

The network output yields a 2048 length floating point vector which we consider as the embedding of an iris image.  To make the feature vector more robust to rotation we employ a simple scheme of concatenating feature vectors based on rotations of the iris image.  Image alignment and rotations are a classic way to improve matching accuracy~\cite{daugman2004iris}. These are done at test time and are called test time augmentations in the literature~\cite{krizhevsky2012imagenet}.  For a single iris image multiple embeddings are calculated by inferring on multiple augmentations of the iris image. These are the original image ($0^\circ$), and then $\{30^\circ, \text{vertical flip, sharpened,horizontal flip,} 330^\circ\}$ One of the augmentation is sharpening the image before inferring the embedding, this is done via an unsharp mask. These augmentations are used in the training process by randomly applying these augmentations to help generalize the network. During testing (embedding calculation) we feed in the original image along with five augmentations of the original image. Ultimately we get six feature vectors per image resulting in a final 12288 length embedding for a single image. This process is visually explained in Figure~\ref{fig:triplet}.

\section{Evaluation}
\subsection{Datasets}
\label{sec:evaluation}
\textbf{ND-IRIS-0405}
The ND-0405~\cite{bowyer2016nd} dataset contains 64,980 iris samples from 356
subjects. This dataset is a superset of the NIST Iris Challenge Evaluation dataset~\cite{phillips2008iris}.  For training, we used the first 25 images of the left irises from all subjects for our training set.  Our test set has 10 randomly chosen images of right irises from all subjects. Recall, that previous studies indicate that the two irises of an individual are statistically independent so in our work right and left iris images of the same person have different class labels~\cite{daugman2004iris}.  Looking ahead, we will evaluate our techniques by training on this dataset and testing on other datasets.  In those experiments, the training dataset is also the first 25 images of each left iris. Images have a resolution of $640\times 480$ while segmented images have a resolution of $256\times 256$. We choose a higher resolution for ND-0405 due to higher resolution of the original images. A lower resolution offers less information and a higher resolution can result in artifacts. We choose an optimal resolution by visually inspecting the segmented images.   We use a deep-learning segmentation system from~\cite{ahmad2018unconstrained} trained using ground truth segmentation from Proenca et al.~\cite{hofbauer2014ground}.

\textbf{IITD}
The IITD dataset~\cite{kumar2010comparison} contains 2,240 image samples
from 224 subjects. The images are have a resolution of $240\times 320$. The segmented images of this dataset are of resolution $200\times 200$. The training set comprises all left eye images while the test set includes five right images selected randomly from all subjects.

\textbf{UBIRIS.v2}
The UBIRIS dataset has 11,102 iris images from 261 different subjects with resolution of $400\times 300$ pixels. This dataset has wide variation among images with off-angle, reflections and imaging distance among the varying parameters introducing some realness to the dataset. The images are also RGB thus do not present clear iris patterns as other NIR datasets. Images in the dataset are taken at different distances which can lead to different results. The segmented images of this dataset are of resolution $200\times 200$.
 All left eye images were used for training while 10 randomly chosen images were taken for testing. As explained above images captured at different distances yield different EER rates, for ensuring unbiased results two test sets of 10 randomly chosen images from testing class are created and the accuracy rates were averaged. 
\subsection{Evaluation scheme}
Feature vectors are generated for every image in the test set. We use an all-all matching scheme where a feature vector is compared with every other feature vector. This is done for all the images in the test set. Large number of images per class can help in the matching process, a feature vector can match closely in its class if it has more templates for its class. This however can also harm the matching process due to the variation when matching across many images. Simple datasets like IITD are ideal for an all-all matching scheme. An unconstrained dataset like Ubiris where there is high variability between images will yield low accuracy in an all-all matching. To compare with the state of the art by~\cite{zhao2017towards} we follow their evaluation scheme. They select the first 10 images of the ND-0405 dataset while we pick 10 images at random to better check the generalizability of the network. Randomly selecting images from the test set yields similar EER numbers so comparison with state of the art is still valid. Some incorrectly segmented images having segmentation inaccuracy of more than 50\% were removed from the testing sets of ND-0405 and Ubiris. We use cosine distance when comparing feature vectors. We find that it delivers better accuracy than using L2 (Euclidean) distance.

\section{Results}
\label{sec:results}

We propose two testing configurations. The \textbf{first} configuration is training and testing on the same dataset. This configuration validates the proposed iris recognition framework and its accuracy on each individual dataset. The hardness of a dataset will skew the results in this configuration. Some level of generalization will be examined since the training and testing sets are completely disjoint. The \textbf{second} configuration has a network trained on the ND-0405 dataset and is used to recognize irises from the remaining datasets. This configuration is designed to check the generalizability of the trained network. We stress that the train/test splits of the datasets are made to compare with the state of the art~\cite{zhao2017towards}.

\begin{figure}[h]
    \centering
    \includegraphics[scale = 0.55]{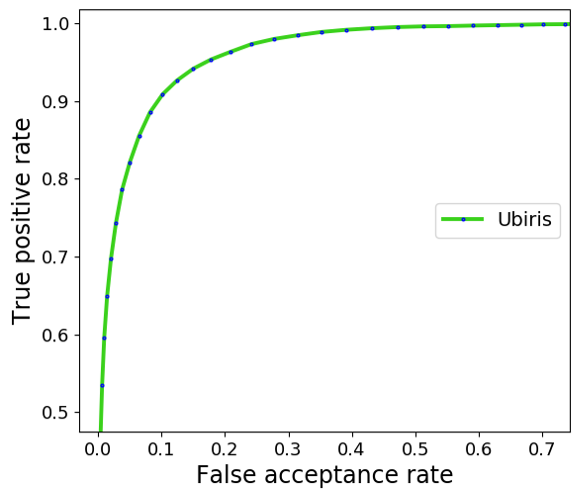}
    \caption{ROC curve for the ubiris dataset.}
    \label{fig:ubiris}
\end{figure}

We randomly selected two testing sets with replacement and report our accuracy rates by averaging the results of the two. We report equal error rate/EER, false reject rate (FRR) at $0.1\%$ false accept rate (FAR) and Rank-1 accuracy. Rank-1 accuracy is how frequently the closest feature vector in the entire test set is in the same class as the image under test.  Results are shown for ND-0405 in Table~\ref{tab:ND-0405}, for Ubiris in Table~\ref{tab:ubiris}, and for IITD in Table~\ref{tab:IITD}.  In these tables we include state of the art prior work for comparison.  In each column, we bold the best system for a particular metric.  We stress that ThirdEye cross is only reported for Ubiris and IITD.  Since it was not trained on the dataset being used for testing, we expect ThirdEye cross to have worse accuracy ThirdEye in all situations.  We show the receiver operator characteristic (ROC) between FAR and FRR for UbirisV2 in Figure~\ref{fig:ubiris} and ND-0405/IITD in Figure~\ref{fig:nd_iitd}.  We note to the reader that these two figures have different axes. 
				
				



\begin{table}[t]
	
		\centering
		\begin{tabular}{|l|r|r|r|}
			\hline
			Method & EER & FRR & Rank-1\\
			\hline
			ThirdEye&1.32\%&8.42\%&\textbf{99.50\%}\\
			\hline
			DeepIrisNet2~\cite{gangwar2019deepirisnet2}&1.47\%&-&-\\
			\hline
			DeepIrisNet2*~\cite{gangwar2019deepirisnet2}&1.48\%&-&-\\
			\hline

			CNN-SVM~\cite{nguyen2018iris}&-&-&98.70\% \\
			\hline
			Imp. Gabor Filters~\cite{li2012iris}&1.70\%&3.73\%&-\\
			\hline
			Triplet network~\cite{zhao2017towards}&\textbf{0.99\%}&\textbf{1.70\%}&-\\
			\hline
	\end{tabular} 
	\caption{Recognition accuracy rates on the ND-0405 dataset.  Dashed items in the table are when a prior work does not report on a metric. Comparison to a network trained on unnormalized images (denoted by *) is also done.}
	\label{tab:ND-0405}

\end{table}

\begin{table}[t]
	
		\centering
		\begin{tabular}{|l|r|r|r|}
			\hline
			Method & EER & FRR & Rank-1\\
			\hline
			ThirdEye&9.20\%&\textbf{60.00\%}&\textbf{83.30\%}\\
			\hline
			DeepIrisNet2~\cite{gangwar2019deepirisnet2}&\textbf{8.50\%}&-&-\\
			\hline
			Zernike moments~\cite{tan2014accurate}&11.96\%&-&63.04\% \\
			\hline
			Imp. Gabor Filters~\cite{li2012iris}&26.14\%&-&-\\
			\hline
			ThirdEye cross&35.00\%&-&25.00\%\\
			\hline

	\end{tabular} 
	\caption{Recognition accuracy rates on the Ubiris.v2 dataset. Dashed items in the table are when a prior work does not report on a metric.}
	\label{tab:ubiris}

\end{table}

\begin{table}[t]
	\centering
		
		\begin{tabular}{|l|r|r|r|}
			\hline
			Method & EER & FRR\\
			\hline
			ThirdEye&\textbf{0.59\%}&0.90\%\\
			\hline
			OSIRIS~\cite{zhao2017towards}&1.11\%&1.61\%\\
			\hline
			Triplet network~\cite{zhao2017towards}&0.64\%&\textbf{0.82\%}\\
			\hline
			ThirdEye cross&1.90\%&0.95\%\\
            \hline

	\end{tabular} 
	\caption{Recognition accuracy rates on IITD. We are not aware of any prior work that reports on Rank-1 accuracy for IITD so it is excluded from comparison.  ThirdEye and ThirdEye cross have a $100\%,99.9\%$ rank-1 accuracy respectively on IITD.}
	\label{tab:IITD}

\end{table}
\subsection{Discussion}
ThirdEye was trained using the methodology presented by Zhao et al.~\cite{zhao2017towards} for comparison purposes. 
As show in Table~\ref{tab:IITD}, on the IITD dataset the EER obtained is slightly better Zhao et al.~\cite{zhao2017towards} but they report a superior  FRR rate at 0.1\% FAR. The IITD is an easy dataset with clear images of the eye coupled with minimum occlusions. We stress that Zhao et al. include a normalization stage in their processing pipeline.  ThirdEye includes no normalization.  Thus, at least on the IITD dataset, accuracy is comparable to state of the art without normalization.

We next turn to the ND-0405 NIR dataset which is the primary dataset used in this work.  The dataset contains some occlusions, motion blur and off-angle images. With the ND-0405 dataset the EER of $1.3\%$ is less than DeepIrisNet2 which again has been trained using normalized images.  DeepIrisNet2 was trained on a much larger training dataset and  tested on a test set slightly smaller than ours. Their work also tries to train networks on unnormalized images in a similar fashion to ours and report EER rates slightly worse than ones from a network trained on normalized images. They also saw networks trained on normalized images converge faster.  We did not observe this phenomenon. 

\begin{figure}[h]
    \centering
    \includegraphics[scale = 0.62]{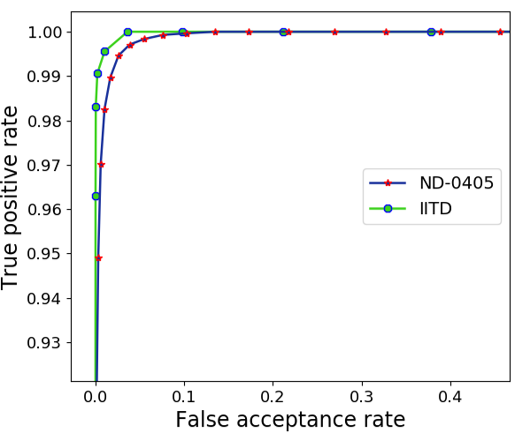}
    \caption{ROC curve for the ND-0405 and IITD datasets.}
    \label{fig:nd_iitd}
\end{figure}
We obtain an EER of 1.3\% which is less than state of the art reported in~\cite{zhao2017towards}. Unlike in the case of IITD, the ND-0405 dataset has occluding eyelashes which we observe are the cause of many false rejects. Our segmentation technique only masks out the iris region leaving occlusions intact.  Zhao et al. use a more robust segmentation technique which removes occlusions.

Ubiris.v2 is the hardest of our testing datasets. The images in visible light do not expose the intricate details of the iris. Ubiris images are also taken at different distances ranging from 4-8 metres. Ubiris is substantially more difficult than the other two datasets.  We do note that the EER reported by DeepIrisNet2~\cite{gangwar2019deepirisnet2} is under different training conditions than our system.  DeepIrisNet2 is trained on a mixture of visual spectrum datasets. Furthermore, they do not specify their matching methodology (we compare all test samples of a class using all to all matching). The particular images used are also not clear. Since images taken from a closer distance are easier to work with on Ubiris.v2. The amount of iris information captured in an image reduces with capturing distance~\cite{proenca2010ubiris}  Despite these differences our ERR is slightly worse than DeepIrisNet2 while our Rank-1 performance is better than state of the art~\cite{tan2014accurate}. 

The ROC curves for ND-0405 and IITD shown in Figure~\ref{fig:nd_iitd} attain a true positive rate of 1.0 at very low FAR values. The major contribution to the high FRR rate for ND-0405 are images with heavy occlusions. Some images in the dataset have eye lashes occluding the iris region, these images are classified correctly with the rank-1 metric however they contribute to the errors in all-all matching. The Ubiris ROC curve in Figure~\ref{fig:ubiris} shows the hardness of the dataset. FRR of 0.0 is reached after an FAR rate of 0.4 signifying that more work is needed to improve recognition on this dataset. Towards that end we use an ensemble network with two ResNets which increases the rank-1 accuracy for ND-0405 to 99.7\%, this however does not change the EER. Furthermore from manual inspection of images and inter-class distances across the testing dataset we observe that the worst inter-class distances occur from iris images which have eyelash occlusions thus leading to our hypothesis that some errors due to segmentation still exist.

\section{Conclusion}
\label{sec:conclusion}
The key question of this work is whether normalization is necessary for accurate iris segmentation.  This work presents a triplet CNN based recognition framework which delivers competitive accuracy performance across the tested datasets. This is despite the fact that it directly uses segmented images as input.  Our performance (relative to state of the art) is best on the most constrained datasets.  This runs counter to Lozej et al.'s~\cite{lozej2019influence} intuition that normalization should hurt recognition more in unconstrained environments. More work is however needed to find the importance of iris normalization in neural network based recognition.

{\small
\bibliographystyle{ieee}
\bibliography{refs}

\begin{thebibliography}{10}\itemsep=-1pt

\bibitem{ahmad2018unconstrained}
S.~Ahmad and B.~Fuller.
\newblock Unconstrained iris segmentation using convolutional neural networks.
\newblock In {\em Proceedings of the Asian Conference on Computer Vision}.
  Springer, 2018.

\bibitem{alonso2009iris}
F.~Alonso-Fernandez, P.~Tome-Gonzalez, V.~Ruiz-Albacete, and J.~Ortega-Garcia.
\newblock Iris recognition based on sift features.
\newblock In {\em 2009 First IEEE International Conference on Biometrics,
  Identity and Security (BIdS)}, pages 1--8. IEEE, 2009.

\bibitem{arsalan2018irisdensenet}
M.~Arsalan, R.~Naqvi, D.~Kim, P.~Nguyen, M.~Owais, and K.~Park.
\newblock Irisdensenet: Robust iris segmentation using densely connected fully
  convolutional networks in the images by visible light and near-infrared light
  camera sensors.
\newblock {\em Sensors}, 18(5):1501, 2018.

\bibitem{bowyer2016nd}
K.~W. Bowyer and P.~J. Flynn.
\newblock The {ND-IRIS}-0405 iris image dataset.
\newblock {\em arXiv preprint arXiv:1606.04853}, 2016.

\bibitem{daugman2004iris}
J.~Daugman.
\newblock Iris recognition border-crossing system in the uae.
\newblock {\em International Airport Review}, 8(2), 2004.

\bibitem{daugman2009iris}
J.~Daugman.
\newblock How iris recognition works.
\newblock In {\em The essential guide to image processing}, pages 715--739.
  Elsevier, 2009.

\bibitem{deng2009imagenet}
J.~Deng, W.~Dong, R.~Socher, L.-J. Li, K.~Li, and L.~Fei-Fei.
\newblock Imagenet: A large-scale hierarchical image database.
\newblock In {\em 2009 IEEE conference on computer vision and pattern
  recognition}, pages 248--255. Ieee, 2009.

\bibitem{gangwar2016deepirisnet}
A.~Gangwar and A.~Joshi.
\newblock Deepirisnet: Deep iris representation with applications in iris
  recognition and cross-sensor iris recognition.
\newblock In {\em 2016 IEEE International Conference on Image Processing
  (ICIP)}, pages 2301--2305. IEEE, 2016.

\bibitem{gangwar2019deepirisnet2}
A.~Gangwar, A.~Joshi, P.~Joshi, and R.~Raghavendra.
\newblock Deepirisnet2: Learning deep-iriscodes from scratch for
  segmentation-robust visible wavelength and near infrared iris recognition.
\newblock {\em arXiv preprint arXiv:1902.05390}, 2019.

\bibitem{he2016deep}
K.~He, X.~Zhang, S.~Ren, and J.~Sun.
\newblock Deep residual learning for image recognition.
\newblock In {\em Proceedings of the IEEE conference on computer vision and
  pattern recognition}, pages 770--778, 2016.

\bibitem{hermans2017defense}
A.~Hermans, L.~Beyer, and B.~Leibe.
\newblock In defense of the triplet loss for person re-identification.
\newblock {\em arXiv preprint arXiv:1703.07737}, 2017.

\bibitem{hofbauer2014ground}
H.~Hofbauer, F.~Alonso-Fernandez, P.~Wild, J.~Bigun, and A.~Uhl.
\newblock A ground truth for iris segmentation.
\newblock In {\em 2014 22nd international conference on pattern recognition},
  pages 527--532. IEEE, 2014.

\bibitem{kerrigan2019iris}
D.~Kerrigan, M.~Trokielewicz, A.~Czajka, and K.~Bowyer.
\newblock Iris recognition with image segmentation employing retrained
  off-the-shelf deep neural networks.
\newblock {\em arXiv preprint arXiv:1901.01028}, 2019.

\bibitem{kinnison2019learning}
J.~Kinnison, M.~Trokielewicz, C.~Carballo, A.~Czajka, and W.~Scheirer.
\newblock Learning-free iris segmentation revisited: A first step toward fast
  volumetric operation over video samples.
\newblock {\em arXiv preprint arXiv:1901.01575}, 2019.

\bibitem{krizhevsky2012imagenet}
A.~Krizhevsky, I.~Sutskever, and G.~E. Hinton.
\newblock Imagenet classification with deep convolutional neural networks.
\newblock In {\em Advances in neural information processing systems}, pages
  1097--1105, 2012.

\bibitem{kumar2010comparison}
A.~Kumar and A.~Passi.
\newblock Comparison and combination of iris matchers for reliable personal
  authentication.
\newblock {\em Pattern recognition}, 43(3):1016--1026, 2010.

\bibitem{lecun2015deep}
Y.~LeCun, Y.~Bengio, and G.~Hinton.
\newblock Deep learning.
\newblock {\em nature}, 521(7553):436, 2015.

\bibitem{li2012iris}
P.~Li and H.~Ma.
\newblock Iris recognition in non-ideal imaging conditions.
\newblock {\em Pattern Recognition Letters}, 33(8):1012--1018, 2012.

\bibitem{liu2016deepiris}
N.~Liu, M.~Zhang, H.~Li, Z.~Sun, and T.~Tan.
\newblock Deepiris: Learning pairwise filter bank for heterogeneous iris
  verification.
\newblock {\em Pattern Recognition Letters}, 82:154--161, 2016.

\bibitem{liu2010finger}
Z.~Liu, Y.~Yin, H.~Wang, S.~Song, and Q.~Li.
\newblock Finger vein recognition with manifold learning.
\newblock {\em Journal of Network and Computer Applications}, 33(3):275--282,
  2010.

\bibitem{lozej2019influence}
J.~Lozej, D.~{\v{S}}tepec, V.~{\v{S}}truc, and P.~Peer.
\newblock Influence of segmentation on deep iris recognition performance.
\newblock In {\em International Workshop on Biometrics and Forensics}, 2019.

\bibitem{ma2004efficient}
L.~Ma, T.~Tan, Y.~Wang, and D.~Zhang.
\newblock Efficient iris recognition by characterizing key local variations.
\newblock {\em IEEE Transactions on image processing}, 13(6):739--750, 2004.

\bibitem{minaee2016experimental}
S.~Minaee, A.~Abdolrashidiy, and Y.~Wang.
\newblock An experimental study of deep convolutional features for iris
  recognition.
\newblock In {\em 2016 IEEE signal processing in medicine and biology symposium
  (SPMB)}, pages 1--6. IEEE, 2016.

\bibitem{nalla2017toward}
P.~R. Nalla and A.~Kumar.
\newblock Toward more accurate iris recognition using cross-spectral matching.
\newblock {\em IEEE transactions on Image processing}, 26(1):208--221, 2017.

\bibitem{nguyen2018iris}
K.~Nguyen, C.~Fookes, A.~Ross, and S.~Sridharan.
\newblock Iris recognition with off-the-shelf cnn features: A deep learning
  perspective.
\newblock {\em IEEE Access}, 6:18848--18855, 2018.

\bibitem{othman2016osiris}
N.~Othman, B.~Dorizzi, and S.~Garcia-Salicetti.
\newblock Osiris: An open source iris recognition software.
\newblock {\em Pattern Recognition Letters}, 82:124--131, 2016.

\bibitem{park2009periocular}
U.~Park, A.~Ross, and A.~K. Jain.
\newblock Periocular biometrics in the visible spectrum: A feasibility study.
\newblock In {\em 2009 IEEE 3rd International Conference on Biometrics: Theory,
  Applications, and Systems}, pages 1--6. IEEE, 2009.

\bibitem{phillips2008iris}
P.~J. Phillips, K.~W. Bowyer, P.~J. Flynn, X.~Liu, and W.~T. Scruggs.
\newblock The iris challenge evaluation 2005.
\newblock In {\em Biometrics: Theory, Applications and Systems, 2008. BTAS
  2008. 2nd IEEE International Conference \ on}, pages 1--8. IEEE, 2008.

\bibitem{proenca2010iris}
H.~Proenca.
\newblock Iris recognition: On the segmentation of degraded images acquired in
  the visible wavelength.
\newblock {\em IEEE Transactions on Pattern Analysis and Machine Intelligence},
  32(8):1502--1516, 2010.

\bibitem{proencca2005ubiris}
H.~Proen{\c{c}}a and L.~A. Alexandre.
\newblock Ubiris: A noisy iris image database.
\newblock In {\em International Conference on Image Analysis and Processing},
  pages 970--977. Springer, 2005.

\bibitem{proenca2010ubiris}
H.~Proenca, S.~Filipe, R.~Santos, J.~Oliveira, and L.~A. Alexandre.
\newblock The ubiris. v2: A database of visible wavelength iris images captured
  on-the-move and at-a-distance.
\newblock {\em IEEE Transactions on Pattern Analysis and Machine Intelligence},
  32(8):1529--1535, 2010.

\bibitem{proencca2018deep}
H.~Proen{\c{c}}a and J.~C. Neves.
\newblock Deep-prwis: Periocular recognition without the iris and sclera using
  deep learning frameworks.
\newblock {\em IEEE Transactions on Information Forensics and Security},
  13(4):888--896, 2018.

\bibitem{raja2014smartphone}
K.~B. Raja, R.~Raghavendra, and C.~Busch.
\newblock Smartphone based robust iris recognition in visible spectrum using
  clustered k-means features.
\newblock In {\em 2014 IEEE Workshop on Biometric Measurements and Systems for
  Security and Medical Applications (BIOMS) Proceedings}, pages 15--21. IEEE,
  2014.

\bibitem{raja2015smartphone}
K.~B. Raja, R.~Raghavendra, V.~K. Vemuri, and C.~Busch.
\newblock Smartphone based visible iris recognition using deep sparse
  filtering.
\newblock {\em Pattern Recognition Letters}, 57:33--42, 2015.

\bibitem{schroff2015facenet}
F.~Schroff, D.~Kalenichenko, and J.~Philbin.
\newblock Facenet: A unified embedding for face recognition and clustering.
\newblock In {\em Proceedings of the IEEE conference on computer vision and
  pattern recognition}, pages 815--823, 2015.

\bibitem{tan2014accurate}
C.-W. Tan and A.~Kumar.
\newblock Accurate iris recognition at a distance using stabilized iris
  encoding and zernike moments phase features.
\newblock {\em IEEE Transactions on Image Processing}, 23(9):3962--3974, 2014.

\bibitem{weinberger2009distance}
K.~Q. Weinberger and L.~K. Saul.
\newblock Distance metric learning for large margin nearest neighbor
  classification.
\newblock {\em Journal of Machine Learning Research}, 10(Feb):207--244, 2009.

\bibitem{wen2016discriminative}
Y.~Wen, K.~Zhang, Z.~Li, and Y.~Qiao.
\newblock A discriminative feature learning approach for deep face recognition.
\newblock In {\em European conference on computer vision}, pages 499--515.
  Springer, 2016.

\bibitem{wildes1997iris}
R.~P. Wildes.
\newblock Iris recognition: an emerging biometric technology.
\newblock {\em Proceedings of the IEEE}, 85(9):1348--1363, 1997.

\bibitem{zhao2017towards}
Z.~Zhao and A.~Kumar.
\newblock Towards more accurate iris recognition using deeply learned spatially
  corresponding features.
\newblock In {\em Proceedings of the IEEE International Conference on Computer
  Vision}, pages 3809--3818, 2017.

\end{thebibliography}
}

\end{document}